\documentclass[letterpaper, 10 pt, conference]{ieeeconf}  

\makeatletter
\let\NAT@parse\undefined
\makeatother

\IEEEoverridecommandlockouts                              

\usepackage{times}  
\usepackage{helvet}  
\usepackage{courier}  
\usepackage{url}  
\usepackage{graphicx}  
\usepackage{color}
\usepackage{amsmath}
\usepackage{amssymb}
\usepackage{multirow}
\usepackage[font={small}]{caption}
\usepackage[utf8]{inputenc}
\usepackage{booktabs}
\usepackage{subfig}
\usepackage{dblfloatfix}
\usepackage{pgfplots}
\usepackage{tikz}
\usepackage{cite}

\usepgfplotslibrary{statistics}
\usetikzlibrary{shapes, arrows}
\usetikzlibrary{backgrounds, fit, calc, 3d}
\usetikzlibrary{positioning}
\pgfplotsset{grid style={dotted,gray}}
\definecolor{dred}{rgb}{1,0,0}
\definecolor{dblue}{rgb}{0,0,1}

\newcommand{\veryshortarrow}[1][3pt]{\mathrel{%
   \hbox{\rule[\dimexpr\fontdimen22\textfont2-.2pt\relax]{#1}{.4pt}}%
   \mkern-4mu\hbox{\usefont{U}{lasy}{m}{n}\symbol{41}}}}

\usepackage{diagbox}

\newcommand\figref[1]{Fig.~{\ref{#1}}}

\newcommand{\Loss}{\mathcal{L}}
\newcommand{\pc}{P}

\newcommand{\ourmodel}{Hind4sight-Net}
\usepackage[hidelinks]{hyperref}
\usepackage{xcolor}
\hypersetup{
	colorlinks,
	linkcolor={red!50!black},
	citecolor={blue!50!black},
	urlcolor={blue!80!black}
}

\title{\LARGE \bf Hindsight for Foresight:  Unsupervised  Structured  \\ Dynamics  Models  from Physical Interaction }

\author{Iman Nematollahi \and Oier Mees \and Lukas Hermann \and Wolfram Burgard \thanks{All  authors  are  with  the  University of Freiburg, Germany. Wolfram Burgard is also with the Toyota Research Institute, Los Altos, USA. This work has  been supported by the German Federal Ministry of Education and Research under contract number 01IS18040B-OML.}} 

\begin{document}
\maketitle
\thispagestyle{empty}
\pagestyle{empty}


\begin{abstract}

A key challenge for an agent learning to interact with the world is to reason about physical properties of objects and to foresee their dynamics under the effect of applied forces.  
In order to scale learning through interaction to many objects and scenes, robots should be able to improve their own performance from real-world experience without requiring human supervision. 
To this end, we propose a novel approach for modeling the dynamics of a robot's interactions directly from unlabeled 3D point clouds and images. Unlike  previous  approaches, our  method does not require ground-truth data associations provided by a tracker or any pre-trained perception network. To learn from unlabeled real-world interaction data, we enforce consistency of estimated 3D clouds, actions and 2D images  with  observed  ones. 
Our joint forward and inverse network learns to segment a scene into salient object parts and predicts their 3D motion under the effect of applied actions. Moreover, our object-centric model outputs action-conditioned 3D scene flow, object masks  and  2D  optical  flow  as  emergent  properties. 
Our extensive evaluation both in simulation and with real-world data demonstrates that our formulation leads to effective, interpretable models that can be used for visuomotor control and planning. Videos, code and dataset are available at \url{http://hind4sight.cs.uni-freiburg.de}

 \end{abstract}

\section{Introduction}

\begin{figure}[t]
	\centering
	\includegraphics[width=1\columnwidth]{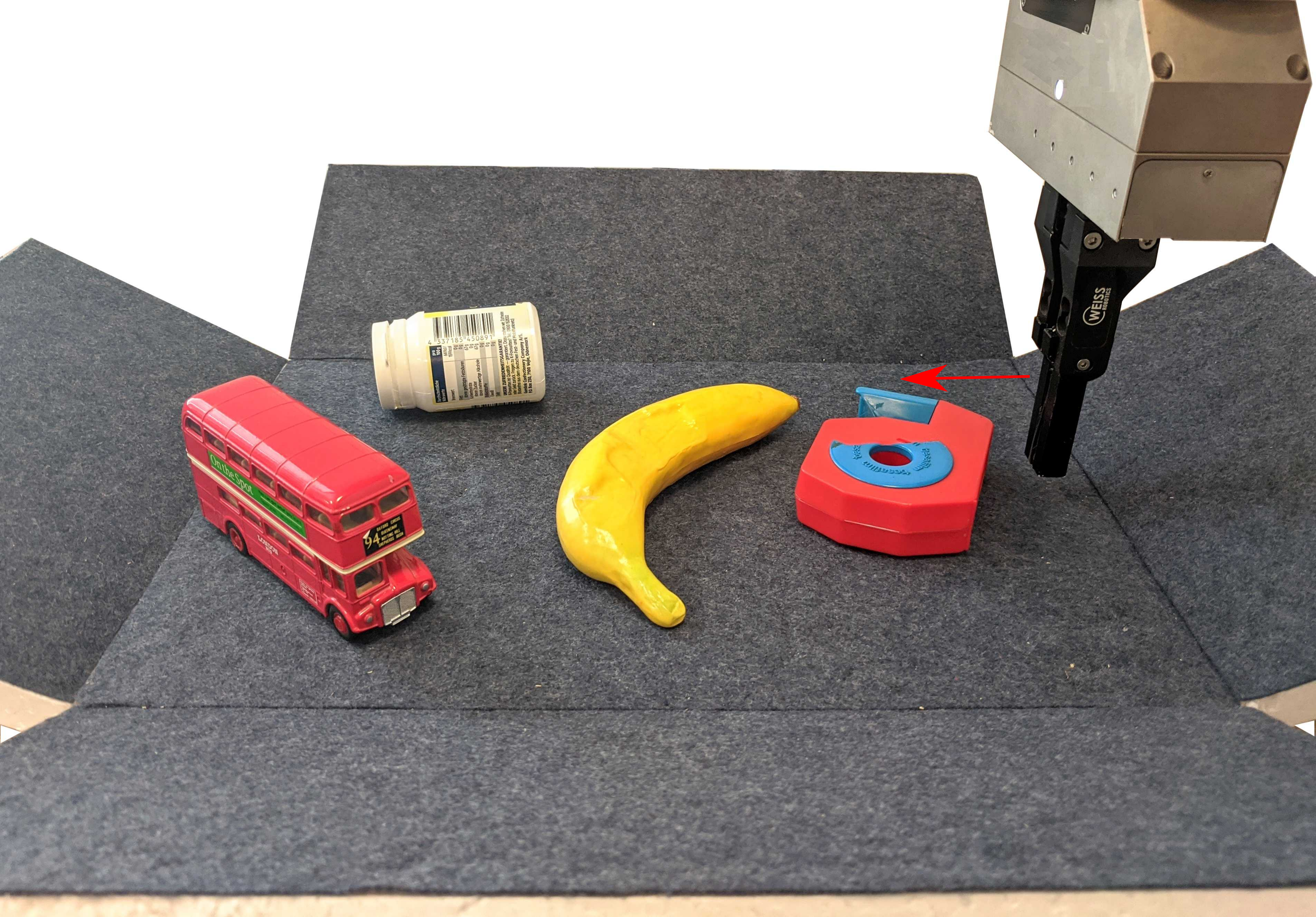}
	\caption{What will happen when the robot arm moves left? Will the tape dispenser collide with the banana? \ourmodel{} learns an unsupervised structured dynamics model which decomposes the scene into objects and predicts their motion conditioned on an action.}
	\label{fig:introduction}
	\vspace*{-4mm}
\end{figure}

What will happen if the robot shown in Figure~\ref{fig:introduction} moves the arm to the left? We can all foresee that the tape dispenser will move to the left, probably colliding with the banana. Intelligent beings have the remarkable ability to effectively interact with unseen objects  by leveraging intuitive models of their environment's physics learned from experience~\cite{battaglia2013simulation, mccloskey1983intuitive}. Predicting the effect of one's actions is a cornerstone of intelligent behavior and also enables reasoning about sequences of actions needed to achieve desired goals. Most existing methods for learning the dynamics of physical interactions are based on high-capacity models, such as deep networks, which can learn complex causal relationships directly from raw sensor data. However, these data-driven methods often suffer from poor sample complexity, requiring large  amounts of data to train and have weaker interpretability and robustness compared to model-based robotics approaches. In contrast, most real-world robot interaction learning methods require human supervision to collect data. Therefore, these
models are trained with small-scale, single-domain data,
leading to reduced generalization capabilities.
Thus, the ability to learn dynamics models autonomously from physical interaction provides an appealing avenue for improving a robot's understanding of its physical environment, as robots can collect virtually unlimited experience through their own exploration.
 
Deep learning has enabled deep predictive models that learn directly in the observation space, relating changes in pixels directly to the applied actions~\cite{ finn2016unsupervised, ebert2018visual, agrawal2016learning}. However, learning to predict physical phenomena from raw video requires handling the high dimensionality of image pixels and discards the knowledge about the structure of the world. Therefore, we explicitly structure our network architecture to decompose the scene into object parts and to predict their dynamics, alleviating the need for predicting pixels.  Our formulation is inspired by SE3-Nets~\cite{byravan2017se3, byravan2017se3posenets}, but relaxes the requirement of ground-truth point-wise data associations. This enables learning scene dynamics in the real-world without external trackers. 

In this paper, we propose a novel approach to learn dynamics of the real-world and present a method that requires neither labeled data nor human supervision, enabling  to improve a robot's understanding of its environment's physics in a lifelong learning manner. Our approach denoted \ourmodel{} jointly learns a forward and an inverse dynamics model and decomposes the scene into salient object parts and predicts their 3D motion. Our  object-centric formulation allows us to capture several desirable inductive biases that help in learning more efficient and interpretable models - a scene comprises of several objects, actions can affect these objects, and the objects can, in turn, affect each other. 
Thus, our network outputs action-conditioned 3D scene flow, object masks and 2D optical flow as emergent properties. We develop a method that combines the flexibility of deep networks with the advantages of model-based approaches, by constraining the learning problem to a low-dimensional interpretable space, as opposed to regressing pixels.   Unlike previous approaches~\cite{byravan2017se3, byravan2017se3posenets, ye2019ocmpc, li2018push, kopicki2017learning, Paus2020}, our method does not require ground-truth point-wise data associations, typically provided by a tracker,  or a pre-trained perception network. 
To learn from unlabeled real-world interaction data, we enforce consistency  of estimated 3D clouds,  actions and 2D images  with  observed  ones.  
Our formulation leads to useful, interpretable models that can be used for visuomotor control and planning. We exemplify this, by using our dynamics model for planning poke actions in both simulation  and with a real robot manipulator.

\section{Related Work}
Our work is primarily concerned with learning intuitive physics~\cite{battaglia2013simulation, mccloskey1983intuitive}. The methodologies to study scene dynamics fall into two paradigms: model-based and data-driven. In order to plan towards a goal state, the model-based approach requires an analytic  physical model of the environment to perform optimal control~\cite{toussaint2009robot}. However,  as many physical properties such as mass and friction cannot be captured easily, assumptions and approximations are often adopted ~\cite{dogar2011framework, rodriguez2012caging, finn2016unsupervised, ebert2018visual, ebert2017self}.

An alternative approach to explicitly modeling the environment via an analytical model is to learn an implicit model of the world using interaction data. There exists a large body of work for understanding intuitive physics from visual cues using deep learning, such as predicting stability of block towers~\cite{lerer2016learning}, learning physic engines~\cite{ battaglia2016interaction,li2019propagation, Paus2020}, estimating object properties~\cite{xu2019densephysnet, zheng2018unsupervised} or object dynamics from images~\cite{mottaghi2016newtonian}. In particular, recent works have looked at mapping raw pixel images to low-dimensional embeddings on top of which standard optimal control methods are applied~\cite{watter2015embed, levine2016end}. In contrast, we use a structured latent representation and predict object masks.
Related to our approach Agrawal \emph{et al.}~\cite{agrawal2016learning} learn a joint forward and inverse model in a feature space where RGB images are encoded, that can be used for poking objects. In comparison, we use an object-centric model that leverages explicit structural constraints and attends to relevant parts of the scene. 
Several works have shown promising results using deep video prediction models for control, either by directly regressing to pixels~\cite{ebert2017self} or using intermediate flow representations~\cite{finn2016unsupervised, ebert2018visual, finn2017deep}. However, these can typically only handle small motions between frames, and need a large number of samples to overcome this inductive bias. 

Our work  addresses learning structured scene dynamics without human supervision, thus falling under the category of self-supervised learning. Due to its ability to learn from unlabeled data, self-supervised learning has been studied in
different sub-fields in AI, such as in computer vision~\cite{doersch2015unsupervised, oord2018representation, vijayanarasimhan2017sfm}, machine learning~\cite{raina2007self} and natural language processing~\cite{devlin2018bert}. Previous works on self-supervised learning in robotics mainly focus on object segmentation~\cite{eitel2019self, pathak2018learning,zeng2017multi}, pose estimation~\cite{deng2019self, mees19iros} or skill learning~\cite{mees20icra_asn, lynch2019learning}. Compared to these approaches, we learn an object-centric structured dynamics model without human supervision.

Most related to our approach is SE3-Nets~\cite{byravan2017se3, byravan2017se3posenets}, a forward model which uses point-wise data associations to approximate  3D rigid object motions for constructing future point clouds. In comparison, our approach is fully unsupervised and therefore enables learning scene dynamics in the real-world without the need of external trackers. To achieve this, our approach adds more explicit structural constraints. Concretely, we force the network to reason over the photometric quality of frame reconstructions resulted from back-projecting the predicted 3D scene flow. Besides combining losses that operate on 3D point clouds and RGB images, we integrate an inverse dynamics model to the network and show that the interplay between both models leads  to  useful,  interpretable  models  that  can  be  used  for  visuomotor control  and  planning.
\begin{figure}[t]
	\vspace*{2mm}
	\centering
	\includegraphics[scale=0.24]{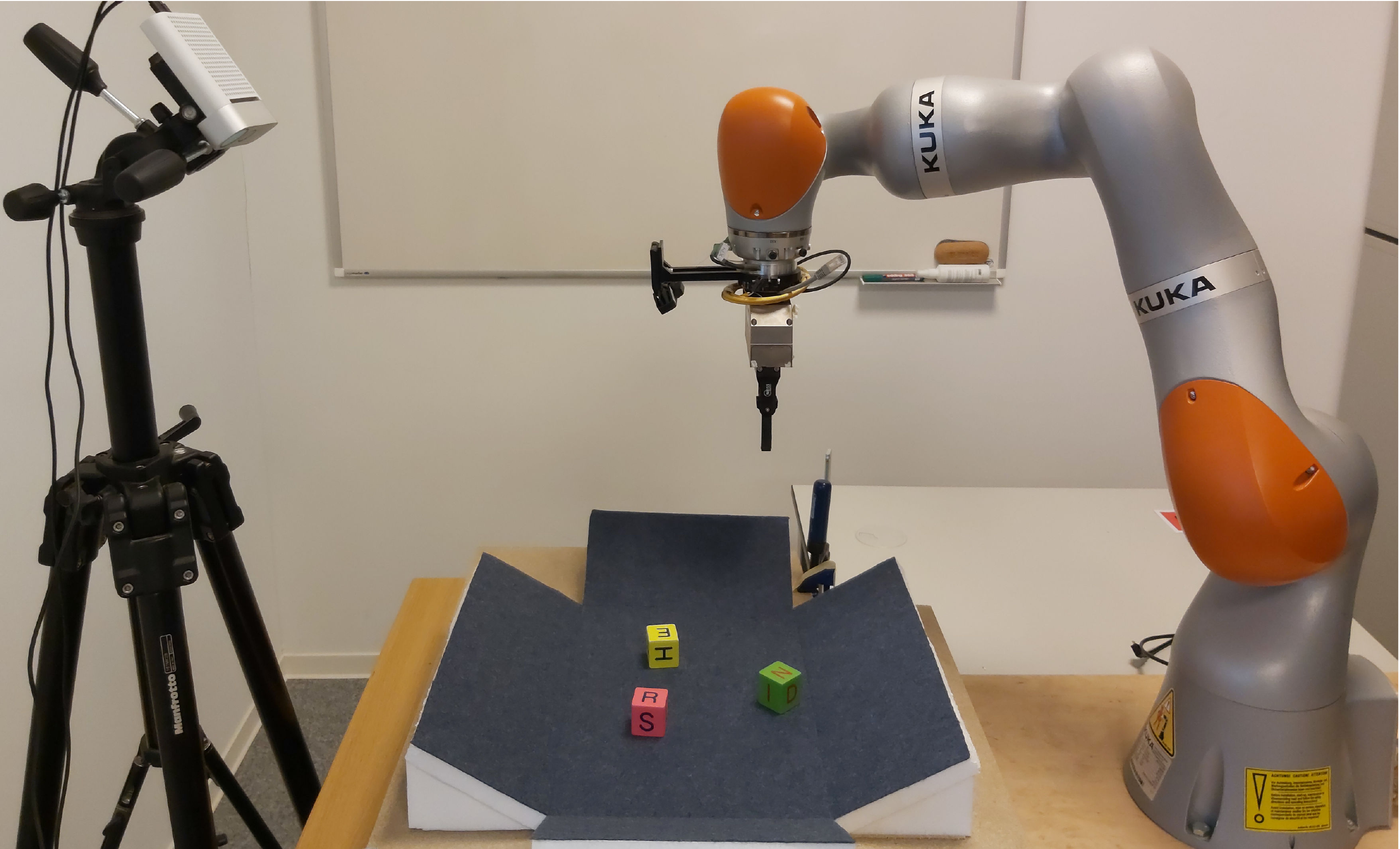}
	\caption{We let a robot interact with objects by randomly poking at them to learn a structured dynamics model. Observational changes in  point clouds and images caused by applied actions constitute the sole learning signals,  enabling  to improve a robot's understanding of its environment's physics in a lifelong learning manner.}
	\label{fig:arena}
\end{figure}
  
  \begin{figure*}[t!]
  	\vspace*{2mm}
	\centering
	\includegraphics[scale=0.3]{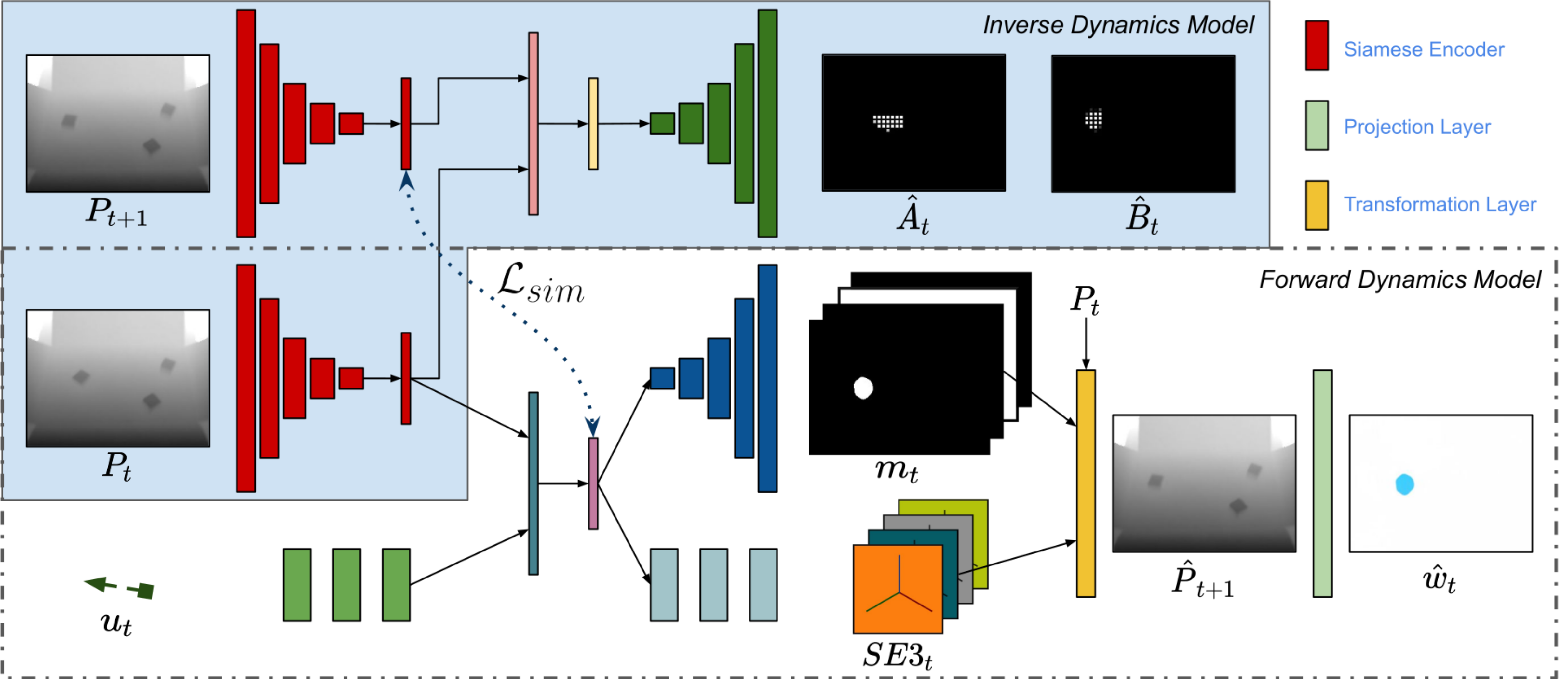}
	\caption{Structure of \ourmodel{}: we jointly learn forward and inverse scene dynamics models from unlabeled interaction data. The forward model segments a 3D point cloud $\pc_t$ of the scene into salient object parts $m_t$ and predicts their $\mathbf{SE}(3)$ motion under the effect of an applied poking action $u_t$. These are then fed into a differentiable ``Transform layer'' that generates the predicted next point cloud $\hat{\pc}_{t+1}$. A ``Projection layer'' back-projects the predicted 3D scene flow into the 2D image plane to retrieve the optical flow $\hat{w}_t$. The inverse model  takes two consecutive 3D point clouds as input and reason over the poking action produced in the form of heat-maps denoting the start $\hat{A}_t$ and end $\hat{B}_t$ point of the poking action.}
	\label{fig:architecture}
\end{figure*}
\section{\ourmodel{}}
In this section we describe the technical details of our
unsupervised structured dynamics model. The architecture of our system is shown in Figure~\ref{fig:architecture}.
Our dynamics model consists of both a forward and an inverse model. A forward model predicts the next world state $\hat{s}_{t+1}$ from the current world state $s_t$ and action $u_t$,  i.e., $\hat{s}_{t+1}=F(s_t, u_t;\theta_{fwd})$, and an inverse model estimates
the action given the initial state and the target state, i.e., $\hat{u}_t=G(s_t, s_{t+1};\theta_{inv})$, where $\theta_{fwd}$ and $\theta_{inv}$ are the parameters of the functions $F$ and $G$.
Predicting which action caused the scene to change is a challenging task for the inverse model, as multiple possible actions can transform the world from one state to another. The inverse model guides the network to construct informative features, which the forward model can then predict and in turn regularize the feature space for the inverse model~\cite{agrawal2016learning}.
Note that in this paper we consider a scenario in which a robot pokes objects on a table and leverages the hindsight from its own interactions to predict dynamics of the scene.

\subsection{Forward Model: Object-centric 3D Motion}
\label{sec:forwardModel}
Our forward model is closely related to SE3-Nets~\cite{byravan2017se3, byravan2017se3posenets}. We take a raw  point cloud $\pc_{t}=(X_t, Y_t, Z_t) $ and an action $u_t$ as inputs and decompose the scene into  $K$ objects, predict their mask $m_{t}^{k}$ and estimate their motion as a 3D rigid body transform $[R,T] \in \mathbf{SE}(3)$ to generate the next point cloud $\hat{\pc}_{t+1}$:
\begin{equation}\label{eq:transform_pointcloud}
\hat{\pc}_{t+1} = \sum_{k=1}^{K} m_{t}^{k} (R_t^{k} \pc_{t} + T_t^{k})
\end{equation}
Note that for points of the scene that lie on the background a mask is assigned as well. Thus, the network learns to attend in which parts of the environment motion occurs. To be more specific, for each point $j$ in the point cloud, $m_{t}^{kj}$ denotes the probability of the point belonging to the $k$-th mask, indicating that each point may be assigned to more than one motion mask. We define the poke action by a poke position and direction. The robot selects a target 2D position $(a_x, a_y)$ on the plane and reaches it from angle $a_\theta$ with respect to the horizon. Hence the poke action vector $u_t$ is a 3-dimensional vector.
Although SE3-Nets showed impressive results, they require ground-truth point-wise data associations as supervision. This means that an external tracking system is needed for acquiring data associations of points in real-world environments. Our approach relaxes this requirement and can be trained without labeled data. 
Concretely, during training we enforce the consistency of estimated 3D clouds, 2D images and actions with observed ones.
\subsection{3D Point Cloud Alignment Loss}
\label{sec:chamferl}
Unlike SE3-Nets that relies on the known data association between the predicted point cloud $\hat{\pc}_{t+1}$ and the target point cloud ${\pc}_{t+1}$ to penalize prediction error, we use the Chamfer distance (CD) between the two points sets to enforce geometric consistency. This distance is a differentiable function that takes as input two points sets ${\pc}_{t+1}$ and $\hat{\pc}_{t+1}$ and for each point in each points set, it finds the nearest neighbor in the other set and sums the squared distances up. Thus, the output of the CD are two continuous distance transforms. We define the distance transforms between the clouds in both directions with $D_{\hat{\pc} \veryshortarrow \pc}^{xy} = \min_{x^{\prime}, y^{\prime}} \|\hat{\pc}_{t+1}^{xy} - \pc_{t+1}^{x^{\prime}y^{\prime}}\|_{2}^2$ and $D_{\pc \veryshortarrow \hat{\pc}}^{xy} = \min_{x^{\prime},y^{\prime}} \|\hat{\pc}_{t+1}^{x^{\prime}y^{\prime}} - \pc_{t+1}^{xy}\|_{2}^2$ and sum them to define the Chamfer distance loss:
\begin{equation}
\label{eq:chamfer_dist}
\Loss_{CD}( \hat{\pc}_{t+1}, \pc_{t+1}) =  \sum_{x,y} \left ( D_{\hat{\pc} \veryshortarrow \pc}^{xy} + D_{\pc \veryshortarrow \hat{\pc}}^{xy} \right )
\end{equation}
\begin{figure*}[t]
	\vspace*{2mm}
	\centering
	\includegraphics[scale=0.24]{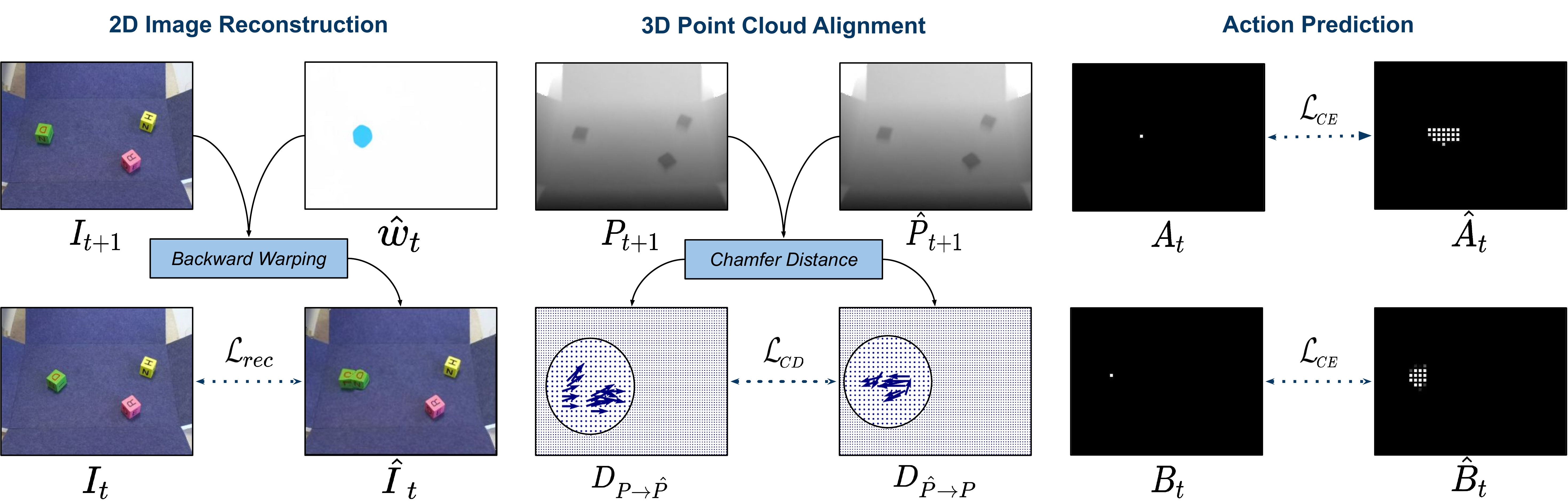}
	\caption{The main loss functions operate on observational changes and enable learning scene dynamics in the real-world without the need of data associations provided by a tracker. The image reconstruction loss uses the predicted 2D flow to minimize a photometric consistency error. The Chamfer Distance tries to enforce the geometric consistency between point clouds. The inverse model predicts spatial distributions of the actions that caused the scene to change.}
	\label{fig:loss_functions}
\end{figure*}
\vspace{-0.75cm}
\subsection{Image Reconstruction Loss}
\label{sec:image_rec}
As learning a dynamics model from scratch without any label or supervision is an ill-posed problem, we reason over the quality of the predicted object motions not only in 3D but also on the image level to better constrain the learning problem. By introducing this constraint, we assume that the brightness of a pixel is not changed by its displacement.
Concretely, we back-project the predicted action-conditioned 3D scene flow into the 2D image plane resulting in 2D optical flow between the two consecutive frames and use backward warping to match pixels from frame $I_{t+1}$ to the frame $I_{t}$ resulting in $\hat{I}_{t}$.
Using known camera intrinsics we project the action-conditioned 3D scene flow into the 2D optical flow  $U_{t}^{xy} = x_{t+1} - x $ and $V_{t}^{xy} = y_{t+1} - y $. Next, we apply a differentiable inverse image warping and minimize the photometric consistency error:
\begin{equation}
\label{eq:photo_loss}
\Loss_{rec}(I_t, \hat{I}_{t}) = \sum_{x, y} \left\|I_t^{xy} - \hat{I}_{t}^{xy}\right\|_{1}
\end{equation} where $\hat{I}_t^{xy} = I_{t+1}^{x^{\prime}y^{\prime}}$ with $x^\prime = x + U_{t}^{xy}$ and $y^\prime = y + V_{t}^{xy}$.
Since image pixels are continuous and back-warped pixels do not always coincide with pixel coordinates, we use a differentiable bilinear sampling~\cite{jaderberg2015spatial} mechanism which interpolates four neighboring pixels of $\hat{I}_{t}^{xy}$ to approximate $I_{t}^{xy}$.
\subsection{Edge-aware Smoothness Loss}
\label{sec:smooth_loss}
In the process of minimizing the photometric consistency error the gradients are mainly derived from the pixel intensity difference between the four neighbors of $\hat{I}_{t}^{xy}$ and $I_{t}^{xy}$. As a consequence, this loss is noisy and would inhibit training if the point is far from the current estimate or located in a low-texture region. Thus, we introduce an edge-aware smoothness loss term to measure the difference between spatially neighbouring points in the flow field, adaptively weighted by the image gradients:
\begin{equation}
\begin{aligned}\label{eq:flow_smooth_loss}
\Loss_{fs} = &\sum_{x, y} |{\nabla U_{t}^{xy}}|  e^{-|\nabla I_{t}^{xy}|} + |{\nabla V_{t}^{xy}}|  e^{-|\nabla I_{t}^{xy}|}
\end{aligned}
\end{equation}
where $| \cdot |$ denotes element-wise absolute value and $\nabla$ is the vector differential operator. 
By enforcing this regularization, we assume that realistic flow fields are piecewise smooth and have discontinuities at the boundaries of moving objects which is a valid assumption for rigid bodies. 

We found it also necessary to apply a similar regularization to the distance transforms computed by the Chamfer Distance. As a nearest neighbor assignment is used to establish a data association between the  predicted and the observed point clouds, this method is prone to spurious matches. Concretely, ``holes'' in the predicted mask for the moving object can be caused by small motions mislead the  nearest neighbor assignment to assume that there is no motion in the overlapping region. We therefore regularize the distance transforms calculated by the CD algorithm in the same manner as our optical flow field to be piecewise smooth:


\begin{equation}
\begin{aligned}\label{eq:dist_smooth_loss}
\Loss_{ds} = &\sum_{x, y} |{\nabla D_{\pc \veryshortarrow \hat{\pc}}^{xy}}|  e^{-|\nabla I_{t+1}^{xy}|} + |{\nabla D_{\hat{\pc} \veryshortarrow \pc}^{xy}}|  e^{-|\nabla I_{t+1}^{xy}|}
\end{aligned}
\end{equation}

\subsection{Inverse Model}
\label{sec:inverse_model}
Our inverse model takes two consecutive raw point clouds $\pc_t$ and $\pc_{t+1}$ as input and predicts the corresponding poke action $\hat{u}_t$ between them through estimating two heatmaps, $\hat{A}_t$ and $\hat{B}_t$, for the start and end positions of the poke respectively. This helps us preserve the similarity of poke actions which take place in close vicinity to each other and ground the actions spatially. To collect training data without human supervision, the robot discretizes the workspace into a grid and marks the two grid cells in which the robot starts and ends its action as keypoints. The objective of the inverse model $\Loss_{inv}$ is a sum of two losses. The term $\Loss_{act}$ is a cross entropy loss on the predicted action heatmaps, while $\Loss_{sim}$ is the L1 loss between the predicted $\hat{s}_{t+1}$ state embedding and the ground-truth $s_{t+1}$ embedding (see Figure~\ref{fig:architecture}). This loss acts as a regularizer between the forward and the inverse model. The inverse model guides the network to construct informative  features,  which  the  forward  model  can  then predict and in turn regularize the feature space for the inverse model.

\subsection{Full Model}
\label{sec:full_model}
Our full model combines all aforementioned objectives to learn a dynamics model from unlabeled data. We abbreviate the losses operating on the 2D image domains to $\Loss_{2D}=\Loss_{rec} + \Loss_{fs}$ and the ones operating on the 3D point clouds as $\Loss_{3D}=\Loss_{CD} + \Loss_{ds}$.
\begin{equation}
\Loss = \lambda_{1}\Loss_{3D}  + \lambda_{2}\Loss_{2D} + \lambda_{3}\Loss_{inv}
\end{equation}

\subsection{Implementation Details}
We adopt a Siamese architecture for the point cloud encoders of our forward and inverse dynamics models. The forward model is based on SE3-Nets~\cite{byravan2017se3} with an additional ``Projection Layer''.
We use ADAM to optimize our model with a learning rate of $10^{-4}$. We weight the main losses with $\lambda_{1} = 10^{5}$, $\lambda_{2} = 10^{3}$ and $\lambda_{3} = 1$.
We train our model for 50 epochs with a mini-batch size of 16. The whole training process of our model is unsupervised and does not need any human annotations. 
We found that initializing one out of $K$ masks to predict all pixels as background and moreover initializing $\mathbf{SE}(3)$ transformations to predict identity transform results in faster convergence. 
\subsection{Model-Predictive Control}
\label{sec:MPC}
Given the learned dynamics model, we can leverage it to find action sequences that lead to a desired goal. We use the cross entropy method (CEM) to search for
the best action sequence~\cite{rubinstein1999cross}, which is a population-based optimization algorithm that infers a distribution over
action sequences that maximize the objective. At every iteration, CEM draws $\mathcal{J}$ trajectories of length $\mathcal{H}$ from a Gaussian distribution, where $\mathcal{H}$ is the planning horizon. We repeatedly evaluate the sampled $\mathcal{J}$ candidate action sequences and re-fit the belief to the top $\mathcal{K}$ action sequences. One advantage of this stochastic optimization procedure is that it allows us to ensure that actions stay within the distribution of actions the model encountered during training. To evaluate a candidate action sequence, we leverage both the 2D and 3D domains our dynamics model has been trained on.

\section{Evaluation}\label{sec:evaluation}
In this section, we evaluate the performance of our unsupervised structured dynamics model on both simulated and real-world datasets and demonstrate its applicability in
a real-world model-predictive control experiment. To the best of our knowledge, there is no publicly available dataset for learning and evaluating dynamics models in the RGB-D domain, as most works consider only RGB images~\cite{agrawal2016learning, finn2016unsupervised, dasari2019robonet}. We therefore evaluate our model on a physics engine and on real interaction data recorded with a robot manipulator.

\begin{table*}[t]
	\vspace*{2mm}
  \centering
  \setlength\tabcolsep{1.8pt}
  \begin{tabular}{ |c|c|c|c|c|c|c|c|c| }
\hline
\multirow{4}{*}{\diagbox{Depth Noise}{Model}} & \multicolumn{3}{c|}{SE3-Nets (Fully-Supervised)} & \multicolumn{4}{c|}{\ourmodel{} (Unsupervised)}  \\ 
\cline{2-8}
& \multicolumn{3}{c|}{DA Noise, threshold = $\pm10$cm} &  \multirow{2}{*}{2D Loss}  & \multirow{2}{*}{3D Loss} &  \multirow{2}{*}{2D Loss + 3D Loss} &  \multirow{2}{*}{Full model}  \\ 
\cline{2-4}
& 0 & $9\times9$ & $11\times11$ &   &  &   &  \\ 
\hline
No Noise &  $1.00 \pm 0.37$ & $1.68 \pm 0.35$ & $2.07 \pm 0.37$ & $1.86 \pm 0.46$ & $1.80 \pm 0.49$ & $1.52 \pm 0.50$ & $\mathbf{1.47 \pm 0.49}$  \\ 
\cline{1-8}
Gaussian Noise, SD = 1cm & $1.07 \pm 0.58$ & $1.94 \pm 0.45$ & $2.20 \pm 0.41$ & $1.93 \pm 0.48$ & $31.39 \pm 24.08$ & $1.78 \pm 0.64$ & $\mathbf{1.63 \pm 0.47}$ \\ 
\hline
  \end{tabular}
  \caption{Average per-point flow MSE in cm under different noise settings for the simulated dataset. Additionally, we analyze the influence of the different losses of \ourmodel{}.}
  \label{tab:ablation}
\end{table*}

\subsection{Poking Task Representation}
\label{sec:smooth_loss}
 
We consider the scenario where a robot is in front of its working arena and a collection of objects lie on top of the arena. The robot collects data by randomly poking objects. The observed scene dynamics are captured with a fixed RGB-D camera. Concretely, before and after each action the depth maps and color images of the scene are stored. 
Random poking can lead to many poke actions being executed in free space, slowing down the data collection of relevant interaction data.
To alleviate this problem, we provide the robot with an observation of the scene without objects and at each interaction perform a background subtraction that discovers actionable parts of the scene. The working arena of the robot is discretized into a 2D grid and at each interaction the robot randomly chooses one occupied cell as the poke target position and one free cell as the poke start position. 
We define the poke action by a target 2D position on the arena and a poke direction $\theta$, corresponding to the angle between start and target cells.

\begin{figure}[t]
	\vspace*{2mm}
	\centering
	\includegraphics[scale=0.13]{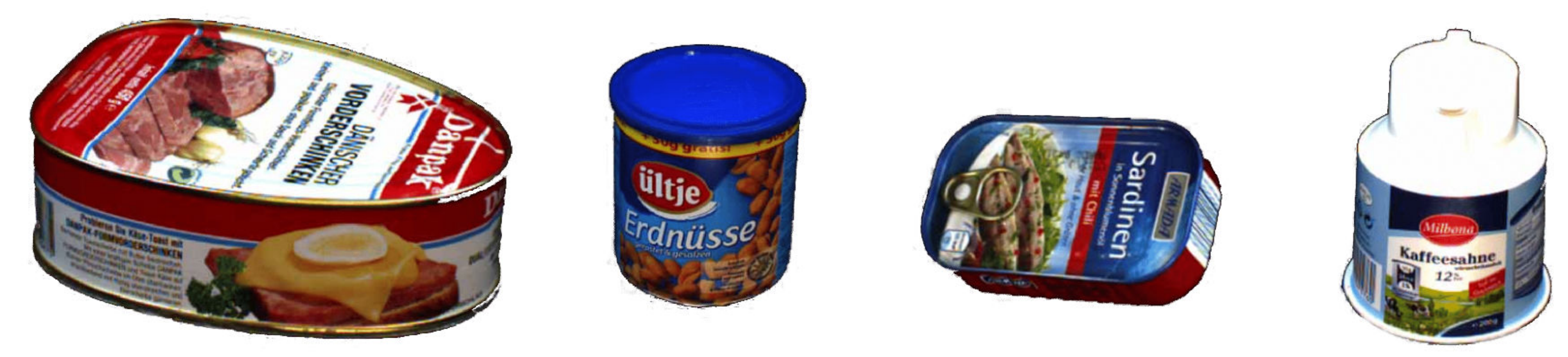}
	\caption{Objects  from the  KIT kitchen object models database~\cite{kasper2012kit}  used in simulation. The objects differ in geometry, size and texture and have varying physical properties.}
	\label{fig:sim_objects}
\end{figure}
 
\subsection{Dataset}\label{sec:sim_dataset}
We evaluate our approach on both synthetic and real data. For experiments on synthetic data, we use the Bullet physics engine~\cite{coumans2016pybullet} to collect a dataset of poking interactions. We pick four representative objects  from the  KIT kitchen object models database~\cite{kasper2012kit}, which differ in geometry, size and texture. These objects are shown in \figref{fig:sim_objects}. 
We record a dataset of 200K interactions, with randomized object start poses and poke actions.  To simulate realistic real-world conditions we also consider noise regarding depth and data association. Concretely, we simulate the noise seen in real depth sensors by adding gaussian noise with a standard deviation (SD) of 1 cm  and scaled the noise by the depth (farther points get more noise). To simulate noise in the data association produced by external tracking systems, we allow for spurious  ground-truth associations. Each point is allowed to be randomly associated to any other point in a $n \times n$ window around it, as long as their depth differences are no larger than $\pm10$cm.

\begin{figure}[t]
\vspace*{2mm}
	\centering
	\includegraphics[scale=0.05]{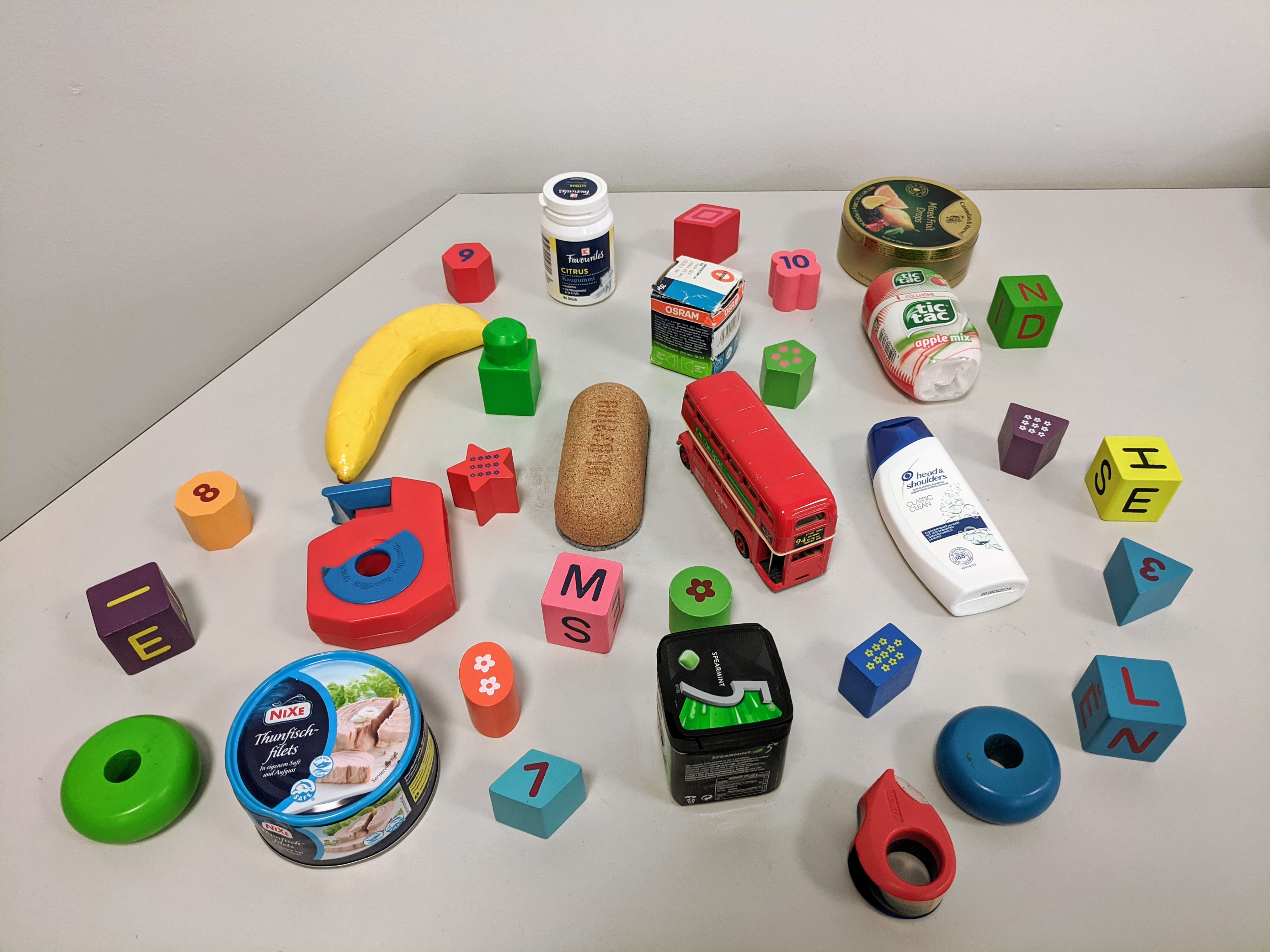}
	\caption{Our real-world poking dataset consists of 34 objects different from each other in shape, appearance, material, mass and friction.}
	\label{fig:all_objects}
\end{figure}

For experiments on real data, we collect 40K of interaction data with a KUKA LBR iiwa manipulator and a fixed Azure Kinect RGB-D camera. We built an arena of styrofoam with walls for preventing objects from falling down. 
At any given time there were 3-7 objects randomly chosen from a set of 34 distinct objects present on the arena. The objects differed from each other in shape, appearance, material, mass and friction as shown in \figref{fig:all_objects}.
Our robot can run autonomously 24/7 without any human intervention, enabling  to  improve a  robot's  understanding  of  its  environments  physics  in  a lifelong learning manner.

\subsection{Evaluation Protocol}\label{sec:evalmetric}

For the quantitative evaluation of the learned structured forward dynamics model, we leverage the Bullet physics engine to access ground-truth action-conditioned scene flow. Following SE3-Nets~\cite{byravan2017se3, byravan2017se3posenets}, we report the Mean Squared Error (MSE) between the predicted 3D scene flow and ground-truth, averaged across points with non-zero ground-truth flow. This metric takes into account errors in both the mask and 3D motion prediction. 

\subsection{Comparisons}\label{sec:comparisons}
The main baseline for our experiments on synthetic data is the fully-supervised SE3-Nets~\cite{byravan2017se3}, as it showed improved performance over SE3-Pose-Nets~\cite{byravan2017se3posenets}. To simulate real-world conditions, we evaluate the performance of SE3-Nets also on moderate settings of noise regarding depth and data association. Thus we evaluate following models:
\begin{itemize}
    \item \emph{SE3-Nets}: The network from~\cite{byravan2017se3} which similarly to us receives a point cloud and an action vector and predicts the next point cloud by decomposing the scene into masks and $\mathbf{SE}(3)$ transformations of attended objects. This model is supervised by the point to point data association of point clouds across two consecutive scenes.
    \item \emph{\ourmodel{}}: Our unsupervised structured dynamics model, which fully exploits available data resources and physically grounded structural constraints by simultaneously learning the forward and inverse models and enforcing the consistency of estimated 3D clouds, actions  and  2D  images  with  observed  ones.
    \item \emph{No motion}: This baseline always predicts zero motion.
\end{itemize}


\subsection{Results on Modeling Scene Dynamics}\label{sec:results_simulation}
We start off by evaluating our method on the scene dynamics recorded with the Bullet physics engine.  To simulate realistic real-world conditions we report our main results on moderate settings of noise regarding depth and data association. To reproduce noise in the data association, we allow for spurious  ground-truth associations  in a $11 \times 11$ window.
Quantitative results of the predicted action-conditioned scene flow are reported in Table~\ref{table:mainResult}. Our \ourmodel{} achieves the best 3D scene flow error compared to baselines even though it fully-unsupervised and not directly trained to predict 3D scene flow. Moreover, our network achieves a large error reduction in comparison to the ``No Motion'' baseline (12.6 cm per point).
\begin{table}[h!]
\centering
\begin{tabular}{ |c|c|c|c } 
\hline
Model & Training Paradigm &  MSE (cm)\\
\hline
SE3-Nets~\cite{byravan2017se3}  & supervised & 2.20 \\ 
\ourmodel{} & unsupervised & $\mathbf{1.63}$ \\ 
No Motion & x & 12.6 \\ 
\hline
\end{tabular}
\caption{Average per-point flow MSE (cm). Our \ourmodel{} achieves the best 3D scene flow error compared to baselines even though it fully-unsupervised and not directly trained to predict 3D scene flow. The ``No Motion'' result quantifies the average magnitude of motion in the dataset.}
\label{table:mainResult}
\end{table}

\begin{figure}[t]
	\centering
	\includegraphics[width=0.9\columnwidth]{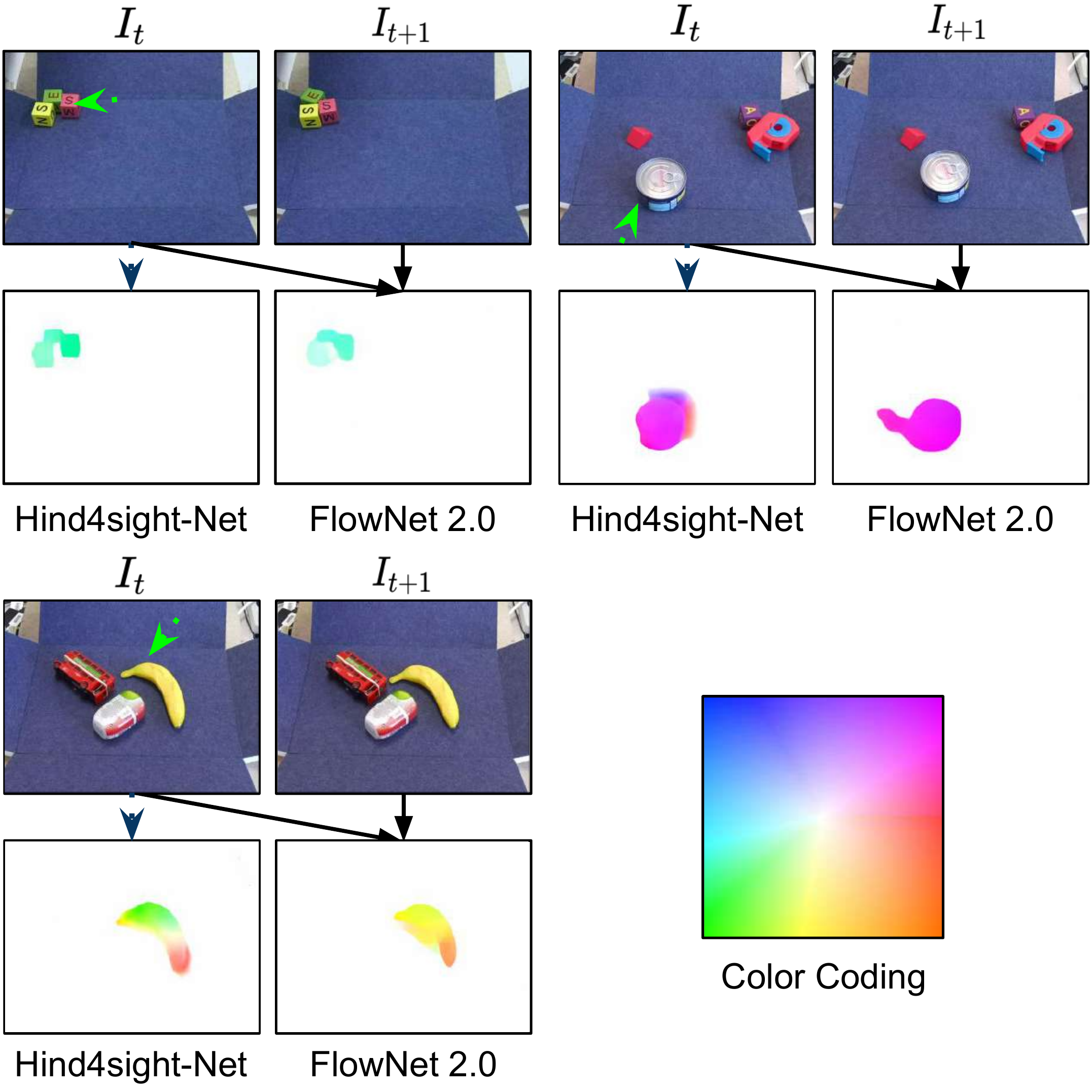}
	\caption{Visualization of the optical flow predicted by FlowNet 2.0~\cite{ilg2017flownet} and the implicit action-conditioned flow learned by our model. \ourmodel{} outperforms FlowNet 2.0
	 as it shows sharper object masks, models collisions better and is less prone to visual distractors such as shadows.}
	\label{fig:realFlow}
\end{figure}

We also evaluate the performance of our implicit action-conditioned 2D optical flow, achieved by projecting the 3D scene flow into the image plane, by comparing it against FlowNet 2.0~\cite{ilg2017flownet}, a state of the art optical flow prediction network. We outperform this strong baseline, despite FlowNet 2.0 having access to two consecutive images as input and having explicit optical flow supervision. Moreover, we observe even better performance for real data, as FlowNet 2.0 is more prone to visual distractors such as shadows (not present in the simulated dataset), see Figure~\ref{fig:realFlow}.
\begin{table}[h!]
\centering
\begin{tabular}{ |c|c|c|c } 
\hline
Model & Inputs & AEE \\
\hline
FlowNet 2.0~\cite{ilg2017flownet} & Images $I_t$ and $I_{t+1}$ & 0.11 \\ 
\ourmodel{} & Point Cloud $\pc_t$ and Action $u_t$& $\mathbf{0.05}$ \\ 
\hline
\end{tabular}
\caption{Average Endpoint Error. Our \ourmodel{} achieves the best 2D optical flow error compared to FlowNet 2.0 even though it is not directly trained to predict 2D optical flow and has no optical flow supervision during training.}
\label{table:AEE}
\end{table}

\begin{figure*}[h]
	\vspace*{2mm}
	\centering
	\includegraphics[scale=0.4]{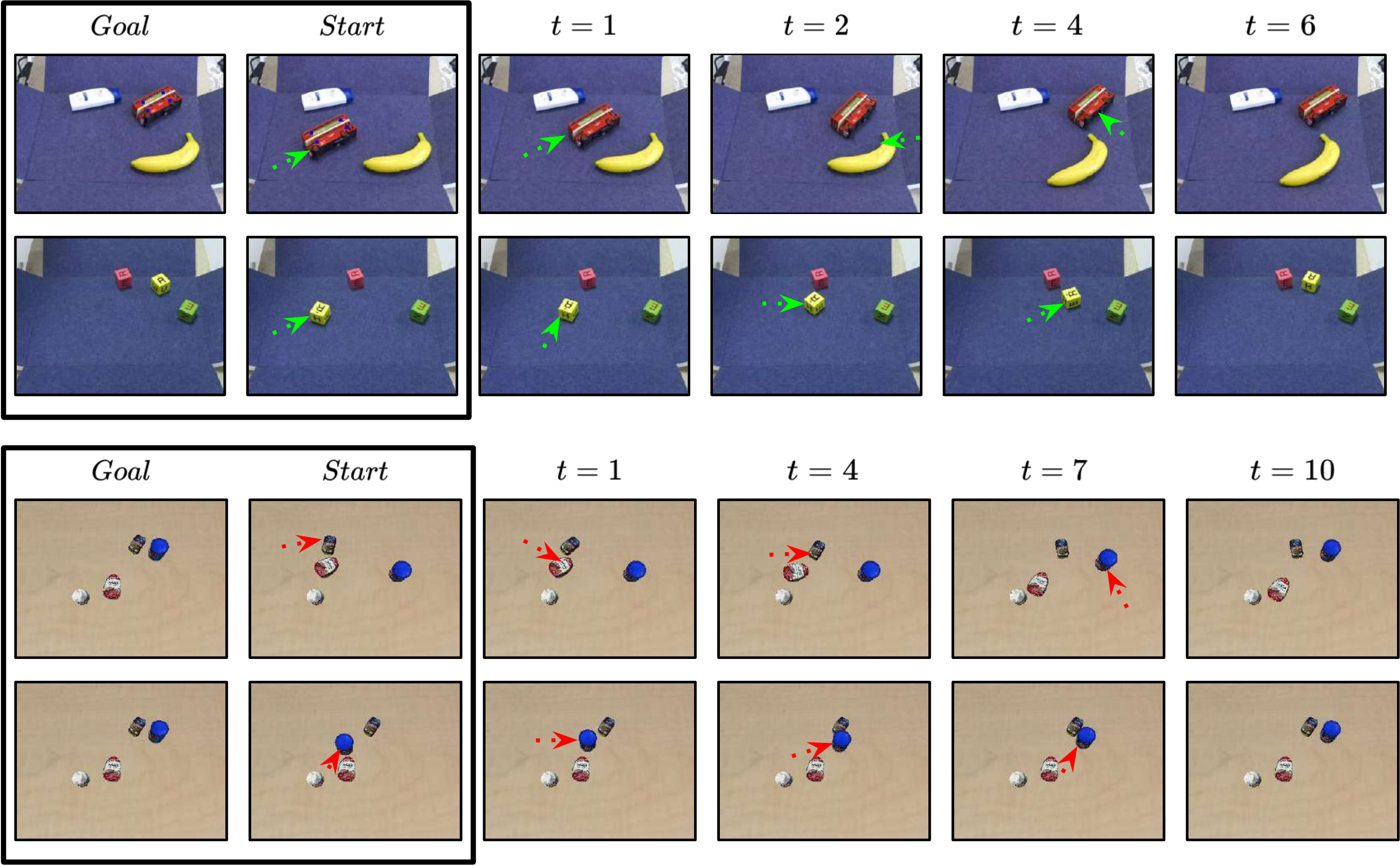}
	\caption{Visualization of executed poking action sequences computed by cross entropy method (CEM) in simulation and real-world: Given the initial configuration and the goal configuration, the arrow shows the sequence of action taken by the robot.}
	\label{fig:planningSteps}
\end{figure*}

\subsection{Ablation Studies}\label{sec:ablation}
To analyze the influence of our different building blocks on the learned dynamics model, we conduct several experiments on the simulated dataset (see Table~\ref{tab:ablation}). Our results indicate that using only a single domain loss is not informative enough to learn an unsupervised dynamics model efficiently. Specially when a realistic noise in the depth is considered, using only the 3D loss leads to large errors due to spurious nearest-neighbor associations. Reasoning jointly over the 3D and image domain improves significantly the results and incorporating the action loss of the inverse model for the full model, achieves the best result. We also evaluate the fully-supervised SE3-Nets under a range of moderate depth and data-association noise conditions and overall observe better performance for our model.


\subsection{Control Performance}\label{sec:results_real}
\begin{figure}[h]
	\centering
	\includegraphics[width=0.9\columnwidth]{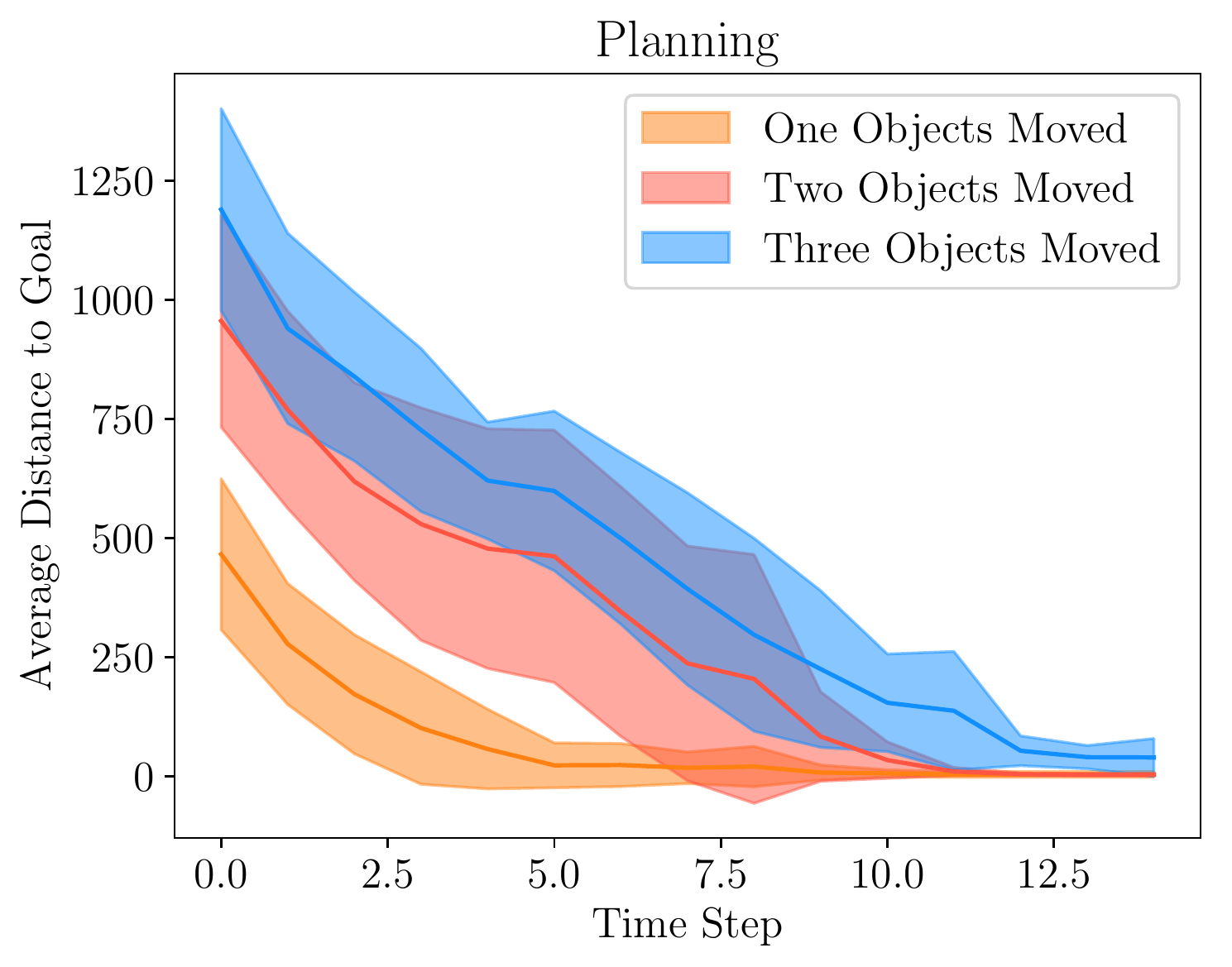}
	\caption{Quantitative results of planning with the learned dynamics model in simulation with variable number of objects to be moved.}
	\label{fig:planning2}
	\vspace*{-4mm}
\end{figure}

To evaluate the effectiveness of the learned dynamics model, we use the cross entropy method (CEM) to find poke action sequences that lead to a desired goal on both simulated and real data.   We define the planning cost-function by a combination of the 3D and 2D domains the network has been trained on. Concretely, we use a combination of the pixel distance, between user marked object points (one point per object) and the Chamfer distance of the whole scene to the goal scene. We use our implicit optical flow to predict how a pixel will move to the next frame given a poke action.
 The pixel distance has a high degree of robustness against distractor objects and clutter, since the optimizer can ignore the values of other pixels. However, we found incorporating global reasoning in 3D space achieved best results, specially to fine-tune the orientation of the target objects. This can be seen as a registration method between a current point cloud and a goal point cloud.
 We perform several experiments by changing the number of objects that need to be moved to reach the goal configuration. We report the average distance of all objects to their respective goal configurations, see Figure~\ref{fig:planning2}. We observe that in most cases we can reach the goal configuration with around 10 poke actions. Moreover, even for the challenging case of planning for three different objects, the learned dynamics model allows to  attend to the relevant parts of the scene and successfully reach the goal configuration, as shown in Figure~\ref{fig:planningSteps}. For simpler tasks with a single object to be moved, we also observe implicit collision-avoidance behavior to some extent, as shown in the last row of Figure~\ref{fig:planningSteps}.
\section{Conclusions and Discussion}
In this paper, we presented a novel approach for learning an ``intuitive'' and structured model of physics from unlabeled robot interaction data. We showed that our formulation enables learning scene dynamics in the real-world without external trackers, human supervision or a pre-trained perception network. We demonstrated that the learned dynamics can be used for visuomotor control and planning. In this work we modeled actions as small pokes, which are likely to be more predictable than large pushing actions. A downside of this choice is that it becomes challenging to observe latent physical properties such as mass and friction from object motion. This is because with a poke action, changes in object movements are mainly  influenced by  dynamics of the manipulator, less so from the object itself. Therefore, investigating an adaptive curriculum learning setup to leverage push actions of variable length and force might be interesting.

Going forward, a natural extension of this work is to try to infer the depth maps directly from the image observations in a self-supervised manner. This would allow to learn structured dynamics model in broader range of applications. Another promising direction for future work is to investigate a tighter coupling of both the inverse and forward dynamics model for planning.

\section*{Acknowledgments}
We would like to thank Arunkumar Byravan and Maxim Tatarchenko for their insightful comments during the development of this work. We further thank Markus Merklinger and Leonhard Sommer for their support while recording the real-world interaction dataset.

\bibliographystyle{IEEEtran}

\footnotesize
\bibliography{literature}

\end{document}